\documentclass{article}

\usepackage[T1]{fontenc}
\usepackage[utf8]{inputenc}
\usepackage{authblk}

\usepackage{times}
\usepackage{soul}
\usepackage{url}
\usepackage[hidelinks]{hyperref}
\usepackage[utf8]{inputenc}
\usepackage[small]{caption}
\usepackage{graphicx}
\usepackage{amsmath}
\usepackage{amsthm}
\usepackage{booktabs}
\usepackage{algorithm}
\usepackage{algorithmic}
\usepackage{multirow}
\usepackage{subfigure}
\usepackage{blindtext}
\usepackage{geometry}
\usepackage{amsfonts}
\usepackage{cite}
\usepackage{subfiles} % Best loaded last in the preamble

\twocolumn  
\usepackage{abstract}

\renewenvironment{abstract}
 {
  \begin{center}
  \Large \bfseries \abstractname\vspace{-.5em}\vspace{0pt}
  \end{center}
  \list{}{
    \setlength{\leftmargin}{.5cm}%
    \setlength{\rightmargin}{\leftmargin}%
  }%
  \item\relax}
 {\endlist}

\title{\Large \textbf{H-OWAN: Multi-distorted Image Restoration with Tensor $1\times1$ Convolution}}

\author[1,2]{Zihao Huang}
\author[2]{Chao Li \thanks{Equal contribution with the first author.}}
\author[1]{Feng Duan}
\author[2]{Qibin Zhao \thanks{Corresponding author. Email:\{chao.li, qibin.zhao\}@riken.jp}}
\affil[1]{College of Artificial intelligence, Nankai University, Tianjin, China}
\affil[2]{Riken Center for Advanced Intelligence Project, Tokyo, Japan}

\date{}
\begin{document}
\maketitle

\begin{abstract}
%\normalsize
It is a challenging task to restore images from their variants with combined distortions.
In the existing works, a promising strategy is to apply parallel ``operations'' to handle different types of distortion. 
However, in the feature fusion phase, a small number of operations would dominate the restoration result due to the features' heterogeneity by different operations.
To this end, we introduce the tensor $1\times 1$ convolutional layer by imposing high-order tensor (outer) product, by which we not only harmonize the heterogeneous features but also take additional non-linearity into account.
To avoid the unacceptable kernel size resulted from the tensor product, we construct the kernels with tensor network decomposition, which is able to convert the exponential growth of the dimension to linear growth.
Armed with the new layer, we propose High-order OWAN for multi-distorted image restoration. 
In the numerical experiments, the proposed net outperforms the previous state-of-the-art and shows promising performance even in more difficult tasks. 
\end{abstract}

\section{Introduction}

Image restoration (IR), the operation of taking a corrupt image and reconstructing its clean counterpart, is a fundamental task in computer vision.
At present, deep-learning-based methods have shown remarkable success in this task particularly when images are corrupted by a specialized type of distortion (\textit{e.g.}, Gaussian noise\cite{tai2017memnet}, Gaussian blur\cite{tao2018scale}).
However, in practical applications like autopilot vision and surveillance, the distortion would be a mixture of various types with unknown strength. 
It therefore degrades the performance of methods in the real world.

\begin{figure}[t] %H为当前位置，!htb为忽略美学标准，htbp为浮动图形
\centering %图片居中
\includegraphics[width=0.5\textwidth]{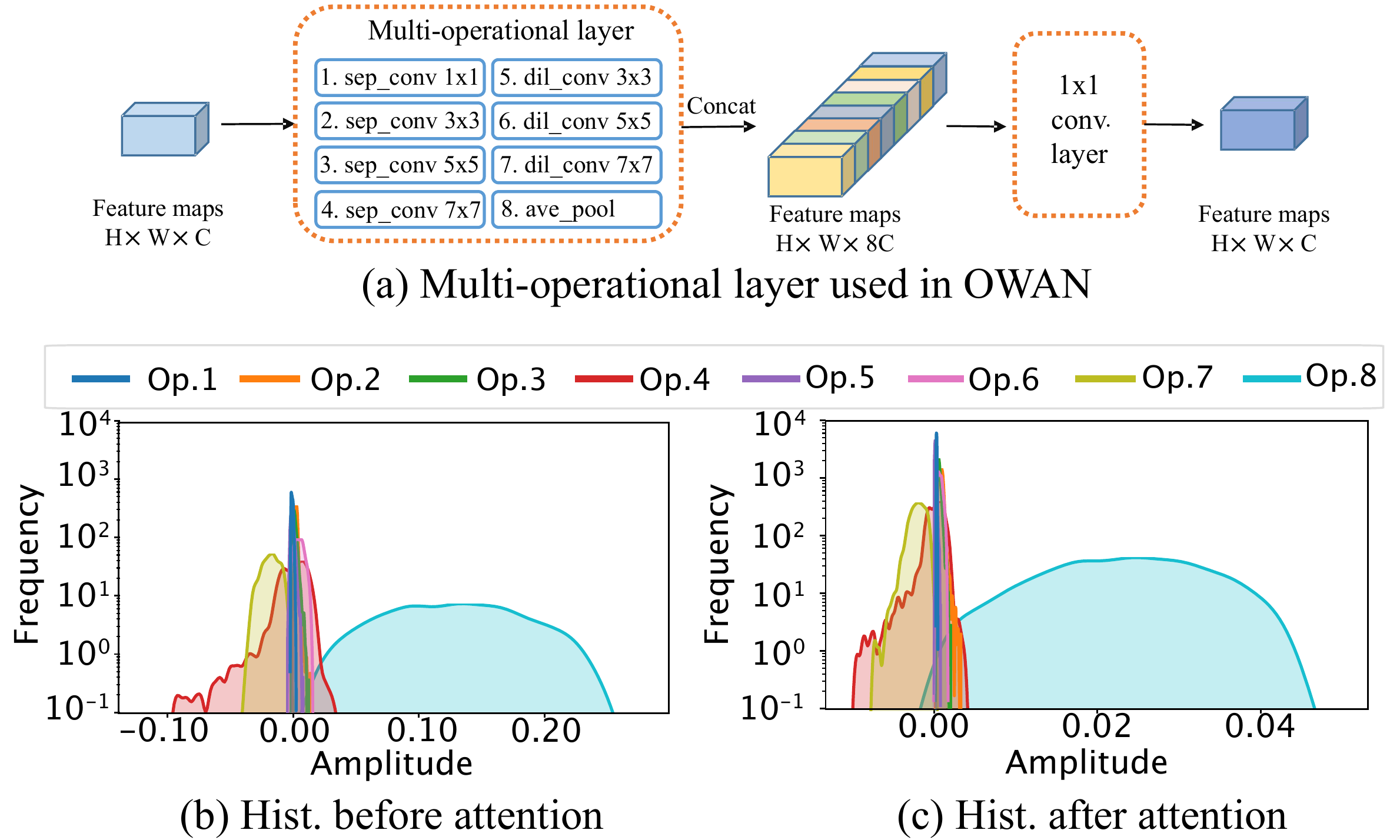} 
\caption{
    Multi-operational layers used in OWAN (a) and the estimated distribution of the feature maps by each operation.
    In Panel (a), the keywords ``sep\_conv'' and ``dil\_conv'' denotes the separable and dilated convolutional layer with specified size, respectively.
    Panel (b) and (c) shows the histogram of the feature maps by eight operations in the test phase, where (b) and (c) corresponds the results before and after the attention module.
    More details are given in Sec. \ref{sec:hetero}.}
\label{fig:1} %用于文内引用的标签
\end{figure}

There are recently several methods proposed to tackle this issue~\cite{yu2018crafting,yu2019path,suganuma2019attention}.
A common idea in these methods is to construct a deep network with multiple ``operational layers/sub-nets'', of which different types are expected to deal with different distortion.
For example, a reinforcement-learning agent is trained in~\cite{yu2019path} for automatically selecting suitable operations.
Operation-wise attention network~(OWAN)~\cite{suganuma2019attention}, as the state-of-the-art (SOTA) approach so far, simultaneously performs eight different operations on feature map following an $1\times{}1$ convolutional layer (see Figure \ref{fig:1} (a)).
Although these methods outperform the previous approaches on the multi-distorted IR task, 
a critical issue is generally omitted in existing methods: \emph{The parallel network architecture with different ``operations'' would lead to heterogeneous feature maps}.
Figure \ref{fig:1} (b) and (c) show the histogram of the feature maps by different operations in OWAN (see Section \ref{sec:hetero} for details).
We will show that some operations would consequently dominate the restoration results due to the heterogeneity.

To this end, we propose a novel tensor $1\times{}1$ convolutional layer (T1CL), by which we can effectively alleviate the aforementioned issue and as the result significantly improve the restoration quality.
Compared to the conventional $1\times{}1$ convolutional layer, the new layer extends the linear operations into multi-linear forms, where for each pixel a tensor kernel is repeatedly multiplied (\textit{i.e.} tensor-product) by the features along every ``direction''~(see Figure \ref{fig:TensorConv}).
Due to the multi-linearity, the entanglement among channels is enhanced.
In the context of the networks like OWAN, concatenating the feature maps by multiple operations along the channel direction, the stronger entanglement is able to harmonize the heterogeneous features and consequently improve the restoration performance.
More interestingly, the experimental results illustrate that the imposed multi-linearity also has the capacity to improve the representation power of the network. 
It implies that the networks equipped with the new layers would achieve promising performance in more challenging tasks.

In Section \ref{sec:hetero}, we discuss the feature heterogeneity and the domination issue in detail by focusing on OWAN.
The notion of tensor $1\times{}1$ convolution layer is introduced in Section \ref{sec:TensorOnebyOneConv},
where we also show \emph{tensor network decomposition}~\cite{oseledets2011tensor,zhao2016tensor} can efficiently reduce the exponentially-increasing dimension of the tensor kernel.

In the experiment, we equip the proposed layer into OWAN by replacing the conventional $1\times{}1$ convolutional layers.
Armed with the new layer, the high-order form of OWAN (\textit{a.k.a.} H-OWAN) outperforms the previous SOTA approaches on the multi-distorted IR task.
Further, the experimental results show that the performance improvement is kept under various hyper-parameters and models.
Last, H-OWAN also shows promising performance in a more challenging task, where more types of distortion are concerned in the experiment.

\subsection{Related Work}

\noindent\textbf{Image restoration} 
% Image restoration (IR) is a fundamental research topic in both image processing and low-level computer vision community~\cite{}.
Given a specialized type of distortion, at present, the state-of-the-art performance is generally achieved by approaches with deep convolutional neural networks (CNNs)~\cite{tai2017memnet,pan2016blind,tao2018scale,kirmemis2018learned,qian2018attentive,lin20182net} to name a few.
% In most of these methods, it is generally assumed that images are corrupted by a specialized distortion type and with known strength.
% Such assumption would lead to performance degeneration since the images we collect in the real world are simultaneously corrupted various types of distortion with unknown strength.
On the other hand, there is few studies focusing on the IR task with combined distortion under unknown strength, \textit{i.e.} multi-distorted image restoration. In a relatively early stage, ``DnCNN''~\cite{zhang2017beyond}, a residual-CNN inspired method, was proposed to deal with blind Gaussian denoising problem. More recently, \cite{yu2018crafting} tackle the multi-distorted IR task using ``RL-Restore'', which learn a policy to select appropriate networks from a ``toolbox''.
Also using reinforcement learning, ``Path-Restore''~\cite{yu2019path} is able to adaptively select an appropriate combination of operations to handle various distortion.
Apart from the methods above, \cite{suganuma2019attention} proposed ``OWAN'', a deep CNN with multi-operational layer and attention mechanism, which achieved the state-of-the-art performance on the multi-distorted IR task.
In contrast to developing novel architectures, in this paper we focus on the heterogeneity and domination issue of feature maps due to the parallel structure of operations/subnets~(especially in OWAN).
We argue that such heterogeneity would degenerate the performance, but this issue can be alleviated by the proposed tensor $1\times{}1$ convolutional layer.

\noindent
\textbf{Feature fusion with tensor product}
Tensor (or outer) product is popularly used in deep learning for feature fusion, and achieves promising performance in various applications.
One line of work is to fuse the features from multi-modal data like visual question answering (VQA)~\cite{ben2017mutan} or sentiment analysis~\cite{liu2018efficient}. 
In these methods, different feature vectors will multiply tensor weights along different directions.
Another line of the work is generally called polynomial/high-order/exponential trick~\cite{hou2019deep,yu2017long,novikov2016exponential}. 
In contrast to the cases in multi-modal learning, the tensor weights are generally symmetric and will be repeatedly multiplied by the same feature vector.
Furthermore, in both two lines, tensor decomposition is generally used for dimension reduction.
The proposed layer in this paper is inspired by the second line of this work.
The difference is that the focus of our work is on the heterogeneity issue rather than multi-modal feature fusion.
Furthermore, to our best knowledge, it is the first time to apply this higher-order structure to the extension of convolutional layers.

\section{Features' Heterogeneity in OWAN}
\label{sec:hetero}
Below, we focus on the OWAN method to discuss how multiple operations lead to heterogeneous feature maps and show that part of the operations would dominate the restoration results in the interference phase.

\begin{figure}[htbp] %H为当前位置，!htb为忽略美学标准，htbp为浮动图形
\centering %图片居中
\includegraphics[width=0.45\textwidth]{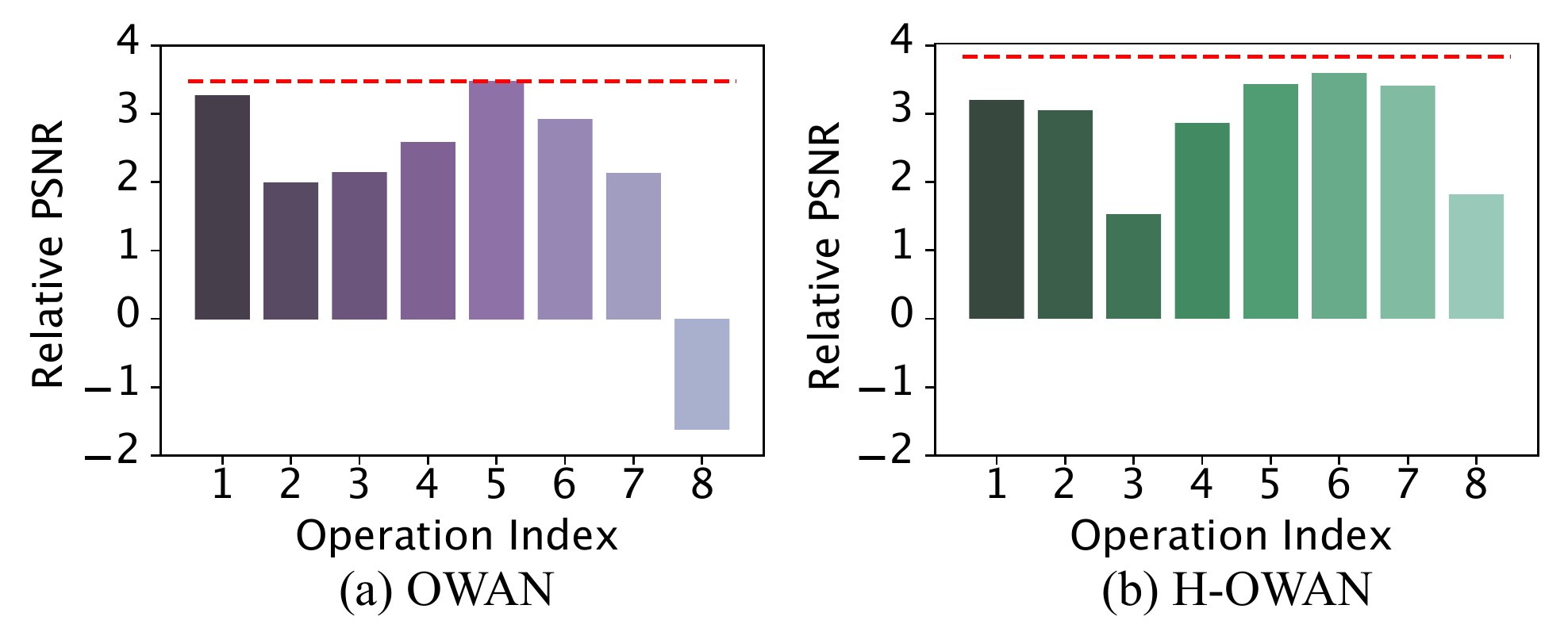} 

\caption{Relative PSNR of the test set when removing one operation. Relative PSNR is the difference with a benchmark, which is the PSNR of the distorted images. The red line of dashes means the performance of maintaining all operational layers.}
\label{fig:HeteroExpResults} %用于文内引用的标签
\end{figure}

Recall the multi-operational layer used in OWAN~\footnote{Here we omit the corresponding attention path for brevity.}.
As shown in Figure \ref{fig:1}, the feature maps are filtered by eight different operations in parallel.
The filtered features are subsequently concatenated and go through a $1\times{}1$ convolutional layer.
To verify the features' heterogeneity from different operations, we set up the following illustrative experiments: For simplicity, we squeeze the scale of OWAN with only 4 multi-operational layers and use randomly selected 5000 and 3584 patches from the dataset for training and testing, respectively.
In the training phase, Adam optimizer is applied until 100 epochs such that the network converges completely.
Panel (b) and (c) in Figure \ref{fig:1} shows the estimated distribution of the features \textit{w.r.t.} each operation of the $4$th multi-operational layer in the inference phase, where the two panels (b) and (c) correspond the results before and after the attention operation, respectively.
We can see that the distributions are significantly different from each operation:
Most of them are quite close to zero, while some are spread in a wide range of values.

The reason resulting in this issue is due to the very different structures of the operations. 
% It is because the operations have different structures.
For example, Op. 8 represents the average pooling layer, of which the output value is naturally larger than ones by convolutional layers with small weights.
Compared between the two plots, the attention module seems to be able to relatively weaken the heterogeneity, but the effect is only on the scale and might not be significant.

Next, we show how much contribution is made by each operation on the restoration task.
To do so, we evaluate the peak relative signal-to-noise ratio (PSNR) of the restored test samples under the condition that we ``close'' the outputs of each operation in turn by setting them to equal $0$.
Figure \ref{fig:HeteroExpResults} (a) shows the experimental results, where the red dashed line represents the performance without closing any operation.
As shown in Figure \ref{fig:HeteroExpResults} (a), the performance is significantly decreased when Op. 8 is closed, while the output by Op. 5 have almost no influence on the performance.
It implies that in OWAN the contribution by different operations is unbalanced. 
Some of the operations like Op. 2 and 8 dominate the restoration results, while some of them like Op. 1 and 5 have little contribution.
Such fact goes against the expectation that the multiple operations play their own to handle various types of distortion.

\noindent\textbf{Can batch-normalization solve this issue?}
One may argue that the heterogeneity could be removed by imposing batch-normalization (BN) layer for each operation.
Note by lots of studies that the restoration quality would be decreased when incorporating BN layers in the network~\cite{liang2017single,pan2018two}.
It is because BN would lead to the interaction of the restored images in the same batch. 
Furthermore, BN can only normalize the 1st and 2nd-order statistical characteristic of the features, and the higher-order characteristics are still out of control.

\section{Tensor $1\times1$ Convolutional Layer}
\label{sec:TensorOnebyOneConv}
In this section, we first mathematically analyze the reason leading to the domination issue.
After that, to address this issue, we propose an extension of the $1\times{}1$ convolutional layer by imposing $p$th-order tensor product, and further introduce how to exploit tensor network decomposition~\cite{oseledets2011tensor,zhao2016tensor} to reduce the unacceptable size of tensor kernels.

\noindent\textbf{Notation}
For brevity of the formulas, we apply the Einstein notation to describe tensor-vector and tensor-tensor multiplication below~\cite{kolda2009tensor}.
For example, assume $x_i$ and $W_{ij}$ to denote a vector and $2$rd-order tensor (\textit{a.k.a.} matrix), respectively, then their product can be simply written as $y_j=W_{ij}x_i$.
Given two vectors $x_i^{(1)}, x_i^{(2)}\in{}\mathbb{R}^C$, we define the concatenation of two vectors as $x_i^{(1)^\frown{}(2)}\in{}\mathbb{R}^{2C}$.
In more general case, the concatenation of $N$ vectors can be simply denoted by $x_i^{(1)^\frown{}(N)}$ without ambiguity. 
Given a vector $x_i\in{}\mathbb{R}^{C}$, the $p$th-order tensor product of $x_i$ is denoted by $\left(x^p\right)_{i_1,\ldots,i_p}\in{}\mathbb{R}^{C^p}$.

\noindent\textbf{Convolution with heterogeneous input}
Assume that we have totally $N$ operations, and given a pixel let  $x_i^{(1)},\,x_i^{(2)},\ldots,x_i^{(N)}$ denote the output feature vectors for each operation, respectively. 
Since in OWAN these outputs are concatenated and subsequently go through a $1\times{}1$ convolutional layer (refer to Figure \ref{fig:1} (a)), the corresponding feature on the output side can be formulized as 
\begin{equation}
    \begin{split}
        y_j&=\phi\left(W_{ij}x_i^{(1)^\frown{}(N)}\right)\\
        &=\phi\left(W_{ij}^{(1)}x_i^{(1)}+\cdots+\underbrace{W_{ij}^{(k)}x_i^{(k)}}_{k\mbox{th-op.}}+\cdots{}+W_{ij}^{(N)}x_i^{(N)}\right)
    \end{split},\label{eq:OnebyOne}
\end{equation}
where $\phi(\,\cdot\,)$ denotes the activation function, $y_j$ denotes the output feature vector given a pixel, and  $W_{ij}$ and $W_{ij}^{(n)},\,n\in{}[N]$ represent the kernel and its partitions \textit{w.r.t.} $x_i^{(n)}$, respectively.
As shown in Equation \eqref{eq:OnebyOne}, the feature $y_j$ can be ``decomposed'' as a sum of components following a non-linear function $\phi$, and we can see each component corresponds to different operations.
It implies that one operation only affects one component in Equation \eqref{eq:OnebyOne}.
It naturally results in the fact that the value of $y_i$ would be dominated if there exist components with a wide range of values (like Op. 4 and 8 in Figure~\ref{fig:1}), while the components concentrating to zero (like Op. 3 in Figure~\ref{fig:1}) will hardly affect the value of $y_j$.
Hence, we claim that the inherent structure of $1\times{}1$ convolutional layer determines the aforementioned domination phenomena.

\begin{figure}[htbp] %H为当前位置，!htb为忽略美学标准，htbp为浮动图形
\centering %图片居中
\includegraphics[width=0.5\textwidth]{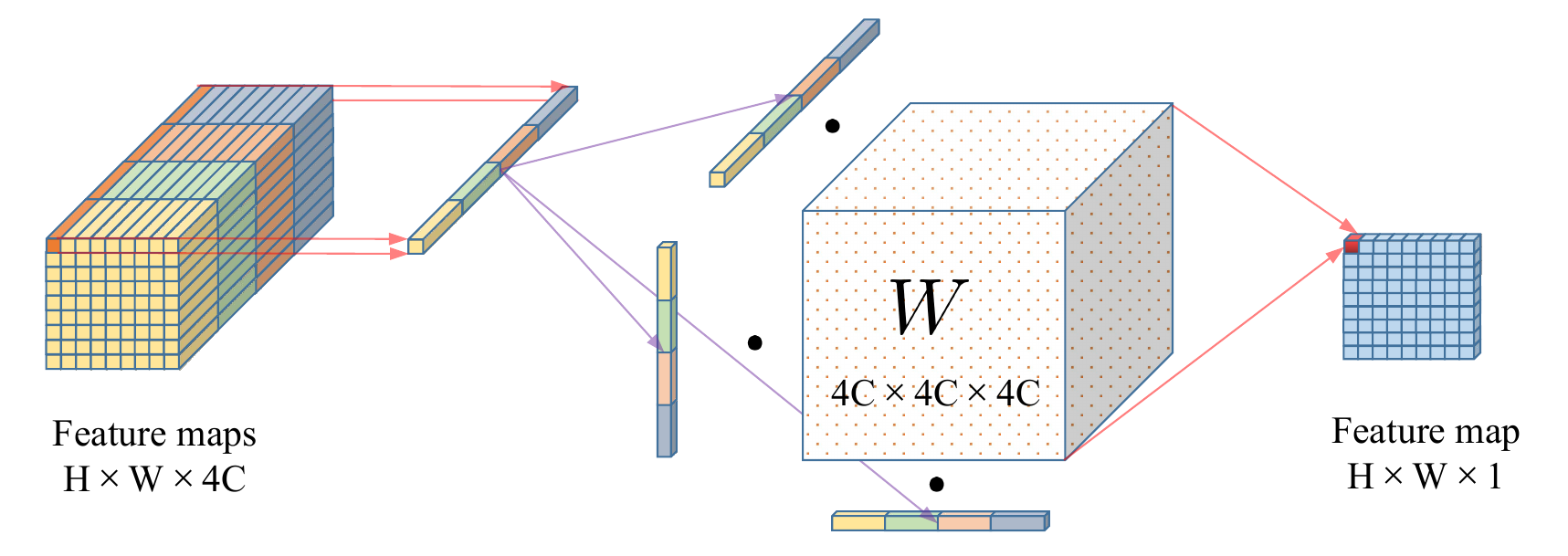} %插入图片，[]中设置图片大小，{}中是图片文件名
% \caption{Illustration of 3-order tensor $1\times1$ convolution. Here we only show the process where the outputs are one channel features. The generalization to multi-channel outputs is similar}
%最终文档中希望显示的图片标题
\caption{Illustration to the $3$th-order tensor $1\times{}1$ convolution, in which the dimension of output channel equals 1.}
% \caption{Illustration of the $3$th-order tensor $1\times1$ convolution, where $\mathcal{W}$ represents the t.}
\label{fig:TensorConv} %用于文内引用的标签
\end{figure}

\noindent\textbf{Convolution via $p$th-order tensor product}
To address this issue, a natural idea is to construct a new form to fuse the features from multiple operations, of which the features can affect as many components in convolution as possible.
Motivated by this, we extend the conventional $1\times{}1$ convolutional layer by imposing $p$th-order tensor product over the feature map.

Specifically,  we extend Equation \eqref{eq:OnebyOne} into a $p$th-order version:
\begin{equation}
    \begin{split}
        y_j&=\phi\left(W_{i_1,\ldots,i_p,j}\left(x^{(1)^\frown{}(N),p}\right)_{i_1,\ldots,i_p}\right)\\
        &=\phi\left(\underbrace{W_{i_1,\ldots,i_p,j}\cdot{}x_{i_1}^{(1)^\frown{}(N)}}_{\mbox{tensor-vec. prod.}}\cdot{}x_{i_2}^{(1)^\frown{}(N)}\cdot{}\cdots{}\cdot{}x_{i_p}^{(1)^\frown{}(N)}\right)
    \end{split}.\label{eq:TensorConv}
\end{equation}
We can see that the tensor kernel $W_{i_1,\ldots,i_p,j}$ is repeatedly multiplied by the same input feature $x_{i_n}^{(1)^\frown{}(N)}$ along $N$ directions.
Figure~\ref{fig:TensorConv} shows an example of the tensor $1\times{}1$ convolution when $p=3$ and $j=1$.
As shown in Figure~\ref{fig:TensorConv}, the kernel is extended into a higher-order tensor compared to the conventional $1\times{}1$ convolutional layer.
% the non-linear property in contrast to 1x1
Also the imposing tensor-product converts the conventional linear convolution into a non(/multi)-linear form.  
% Furthermore, it will be shown in the experiments that the additional non-linearity due to the tensor product would strength the representation power of the network.
The conventional $1\times{}1$ convolutional layer is a special case of the proposed tensor layer when $p=1$.

\begin{figure}[htbp]
\centering    %居中
         %子图居中
\includegraphics[scale=0.3]{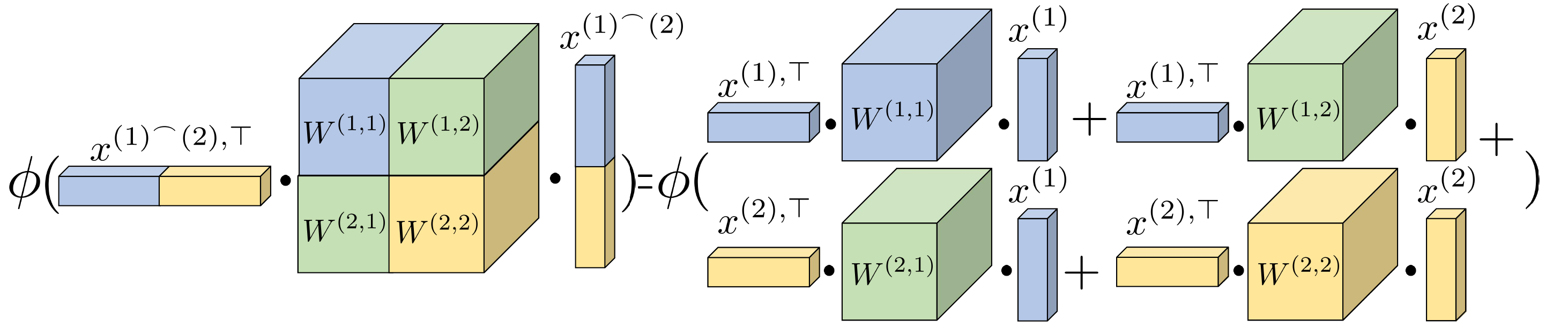}   %以pic.jpg的0.5倍大小输出

\caption{Illustration to Equation (3) with ``decomposition'' \textit{w.r.t.} operations.} %  %大图名称
\label{fig:tensor}  %图片引用标记
\end{figure}

% why the high-order version can address the aforementioned issue.
Next, we show how the tensor $1\times{}1$ convlutional layer solve the aforementioned domination problem.
As an illustrative example, we assume that only 2 operations are concerned and the order of layer $p=2$.
Like Equation \eqref{eq:OnebyOne}, we can also ``decompose'' Equation \eqref{eq:TensorConv} as
\begin{equation}
    \begin{split}
        y_j&=\phi\left(W_{i_1i_2j}\cdot{}x_{i_1}^{(1)^\frown{}(2)}\cdot{}x_{i_2}^{(1)^\frown{}(2)}\right)\\
        &=\phi\left(
        \begin{matrix}
            W_{i_1i_2j}^{(1,1)}\cdot{}x_{i_1}^{(1)}\cdot{}x_{i_2}^{(1)}
            +
            W_{i_1i_2j}^{(1,2)}\cdot{}x_{i_1}^{(1)}\cdot{}x_{i_2}^{(2)}\\
            +W_{i_1i_2j}^{(2,1)}\cdot{}x_{i_1}^{(2)}\cdot{}x_{i_2}^{(1)}
            +W_{i_1i_2j}^{(2,2)}\cdot{}x_{i_1}^{(2)}\cdot{}x_{i_2}^{(2)}
        \end{matrix}
        \right)
    \end{split}.
    \label{eq:ExampleTesnorOnebyOne}
\end{equation}
An graphical illustration of this equation is shown in Figure~\ref{fig:tensor}.
We can see that the tensor product results in more entanglement among different operations.
It implies that,  with increasing the order $p$, the feature vector associated with a given operation would affect more components compared to Equation \eqref{eq:OnebyOne}.
Such entanglement of operations would balance the contribution of the features even though there is a heterogeneous structure among them.

To validate this claim, we re-implement the experiment in Section \ref{sec:hetero} except replacing the conventional $1\times{}1$ convolutional layers by the proposed tensor form with $p=2$.
The experimental results are shown in Figure \ref{fig:HeteroExpResults} (b).
Compared to the results in Figure \ref{fig:HeteroExpResults} (a), we can see that the influence on the restoration quality by each operation is significantly alleviated.

\noindent\textbf{Dimension reduction by tensor network decomposition}

% There are many methods to conduct tensor decomposition such as CP (Canonical Polyadic), TT (Tensor Train), TR (Tensor Ring). By assuming that $\mathcal{W}$ admits some tensor decomposition format, we can get the decomposed core weights directly by learning, which can be directly reduced to the weight tensor in the network inference stage. Here we show 3 types of decomposition format:
% When we use these decomposition method to 
% The green bars in Fig.~\ref{OWAN_heterogeneity} show the results by using OWAN model armed by tensor $1\times1$ convolution, which we call HOWAN, at the same experimental configuration. Compared with OWAN, there is no operation in HOWAN to dominate the output of the model or have no effect on the model.
A critical issue brought from the new layer is that the kernel size will exponentially increased with the order $p$.
To solve this problem, we apply \emph{tensor network (TN)} decomposition to reduce the dimension of the kernel.
TN decomposition is to represent a high-order tensor by a collection of smaller core tensors~\cite{singh2010tensor}.
In this paper, we consider three popular types of TN decompositon models including canoncical/polyadic (CP)~\cite{kolda2009tensor}, tensor-train (TT)~\cite{oseledets2011tensor} and TR~\cite{zhao2016tensor}.
Using the three models, the kernel in a $p$th-order tensor $1\times{}1$ convolutional layer can represented by

\begin{align}
    &W_{i_1,\ldots,i_p,j}^{CP}=G_{i_1,r,j}^{(1)}G_{i_2,r,j}^{(2)}\cdots{}G_{i_p,r,j}^{(p)},\tag{4-CP}\\
    &W_{i_1,\ldots,i_p,j}^{TT}=G^{(1)}_{i_1,r_1,j}\cdots{}G^{(k)}_{i_1,r_{k-1},r_{k},j}\cdots{}G^{(p)}_{i_1,r_{p-1},j}\tag{4-TT},\\
    \mbox{and}\nonumber\\
    &W_{i_1,\ldots,i_p,j}^{TR}=G^{(1)}_{i_1,r_0,r_1,j}\cdots{}G^{(k)}_{i_1,r_{k-1},r_{k},j}\cdots{}G^{(p)}_{i_1,r_{p-1},r_0,j},\tag{4-TR}
\end{align}
respectively. 
In the equations, the internal indices $r_k,\forall{}k$ is usually called bound dimension in physics literature~\cite{gerstner2003dimension} or \emph{rank} in computer science~\cite{liu2018efficient}, which controls the total number of parameters used in the layer.
Since in layers the tensor kernel is multiplied by the same vector along all but the channel directions.
Hence it is naturally to further assume the symmetric structure of the kernel, \textit{e.g.,} $G_{i_k,r,j}^{(k)}=G_{i_l,r,j}^{(l)}\,k\neq{}l$ for the CP decomposition or $G_{i_k,r_{k-1},r_k,j}^{(k)}=G_{i_l,r_{l-1},r_{l},j}^{(l)}\,k\neq{}l$ for TR.

\noindent\textbf{Complexity analysis}
Assume that the dimension of the input and output feature vectors to be equal to $I$ and $J$, respectively. 
In this case, for each sample both the computational and storage complexity of the conventional $1\times{}1$ conventional layer equals $\mathcal{O}\left(IJ\right)$ per pixel, while it increases to $\mathcal{O}\left(I^pJ\right)$ for the vanilla $p$th-order form.
If the kernel is represented by TN decomposition, the complexity can be decreased to $\mathcal{O}\left(pRIJ\right)$ for rank-$R$ CP, and computationally $\mathcal{O}\left(p(R^2I+R^3)J\right)$ and storely $\mathcal{O}\left(pR^2IJ\right)$ for both TT and TR models with rank-$R$.
We can see that TN decomposition successfully convert the complexity from exponential growth to a linear growth associated with the order $p$.
In practice, the value of rank $R$ is generally small, thus TN decomposition can significantly reduce the computational and storage requirement to implement the new layer.

\begin{table*}[ht]
\centering
\begin{tabular}{c|cc|cc|cc}
\toprule
Test set   & \multicolumn{2}{c|}{Mild (unseen)} & \multicolumn{2}{c|}{Moderate}    & \multicolumn{2}{c}{Severe (unseen)} \\ \hline
Metric     & PSNR            & SSIM             & PSNR           & SSIM            & PSNR             & SSIM             \\ \hline
DnCNN     &    27.51       &     0.7315        &     26.50     &   0.6650        &    25.26       &     0.5974       \\
RL-Restore & 28.04           & 0.7313           & 26.45          & 0.6557          & 25.20            & 0.5915           \\
Path-Restore       & N/A           & N/A           & 26.48          & 0.6667          & N/A            & N/A           \\
OWAN     & 28.33  & 0.7455           & 27.07 & 0.6787          & 25.88   & 0.6167           \\\hline \hline
H-OWAN-order2 (ours)      & 28.39           & \textbf{0.7485}  & \textbf{27.13}          & \textbf{0.6820} & \textbf{26.03}            & 0.6207  \\
H-OWAN-order3(ours)     & 28.35  & 0.7464           & 27.07 & 0.6810          & 25.99   & \textbf{0.6210}           \\
H-OWAN-order4(ours)     & 28.32  & 0.7430           & 27.06 & 0.6791          & 25.92   & 0.6176           \\
H-OWAN-order2-add1(ours)     & 28.40  & 0.7474           & 27.10 & 0.6811          & 25.96   & 0.6183           \\
H-OWAN-order3-add1(ours)     & \textbf{28.43}  & 0.7479           & 27.11 & 0.6814          & 26.00   & 0.6187           \\
H-OWAN-order4-add1(ours)     & 28.38  & 0.7473           & 27.13 & 0.6809          & 26.00   & 0.6185           \\
\bottomrule
\end{tabular}

\caption{
\textbf{Results on DIV2K.}
Comparison of DnCNN, RL-Restore, Path-Restore, OWAN and our models using DIV2K test set. ``order$N$'' denotes the order of the proposed T1CL, and ``add1''  means concatenating the feature map with a constant of 1 at the end.
``N/A'' means no results published.
}
\label{Table1:3 noise compare}
\end{table*}

\section{Experiments}

Below, we design three sets of experiments each addressing a different research question:

\begin{itemize}
    \item [\textbf{E1:}] Armed with the tensor $1\times{}1$ convolutional layers (T1CLs), we compare the modified OWAN (high-order OWAN, H-OWAN) with previous state-of-the-art (SOTA) approaches in multi-distorted image restoration.
    \item [\textbf{E2:}] We study the impact of the hyper-parameters imposed by the new layer like order and rank.
    \item [\textbf{E3:}] We explore whether higher-order layers perform better on more difficult multi-distorted image restoration tasks.
\end{itemize}

\subsection{E1: Comparison with SOTAs}

\noindent\textbf{Network setup}
To demonstrate the effectiveness of the proposed layer,  we follow the same network setup to OWAN except that the used $1\times{}1$ convolutional layers are replaced by the new layers.
The details of H-OWAN are as follows: we set up the network with 10 OWAN blocks~\cite{suganuma2019attention}, each of which contains 4 proposed T1CLs.
For each T1CL, we apply the rank-16 CP decomposition to dimension reduction with the symmetric structures, \textit{i.e.} shared core tensors. 
In the training phase, we apply the batch-averaged $l_1$-distance between the restored images and its groundtruth as the loss function, and Adam optimizer~\cite{kingma2014adam} to training where $\alpha =0.001$, $\beta _{1}=0.9$, and $\beta _{2}=0.99$. 
The initial learning rate equals 0.001 and the cosine annealing technique~\cite{loshchilov2016sgdr} is employed for adjusting. 
And our network is trained by 100 epochs with mini-batch size equaling 32.

\noindent\textbf{DIV2K Dataset}
We evaluate the performance of our network by DIV2K dataset, which is also used in \cite{suganuma2019attention,yu2019path,yu2018crafting}.
% The dataset contain 800 high resolution images for training and 100 same quality images for validation and testing, respectively.
In the experiment, 800 images from DIV2K are selected and divided into two parts: 750 images as the training set and 50 images as the testing set.
In addition, we clip each image into many $63\times63$ patches, where we totally have 230,080 and 3,584 patches in the training and test set, respectively.

Three types of distortion are considered in the experiment including Gaussian noise, Gaussian blur, and JPEG compression. They are mixed and added to the data with a wide range of degradation levels, which are separated into three groups: mild, moderate, and severe levels. 
To simulate the situation of unknown distortion strength,  we only employ the moderate level on training data, but the testing data is generated at all three levels. 
% More details see the experimental section in~\cite{suganuma2019attention}.

\begin{figure}[h] %H为当前位置，!htb为忽略美学标准，htbp为浮动图形
\centering %图片居中
\includegraphics[width=0.45\textwidth]{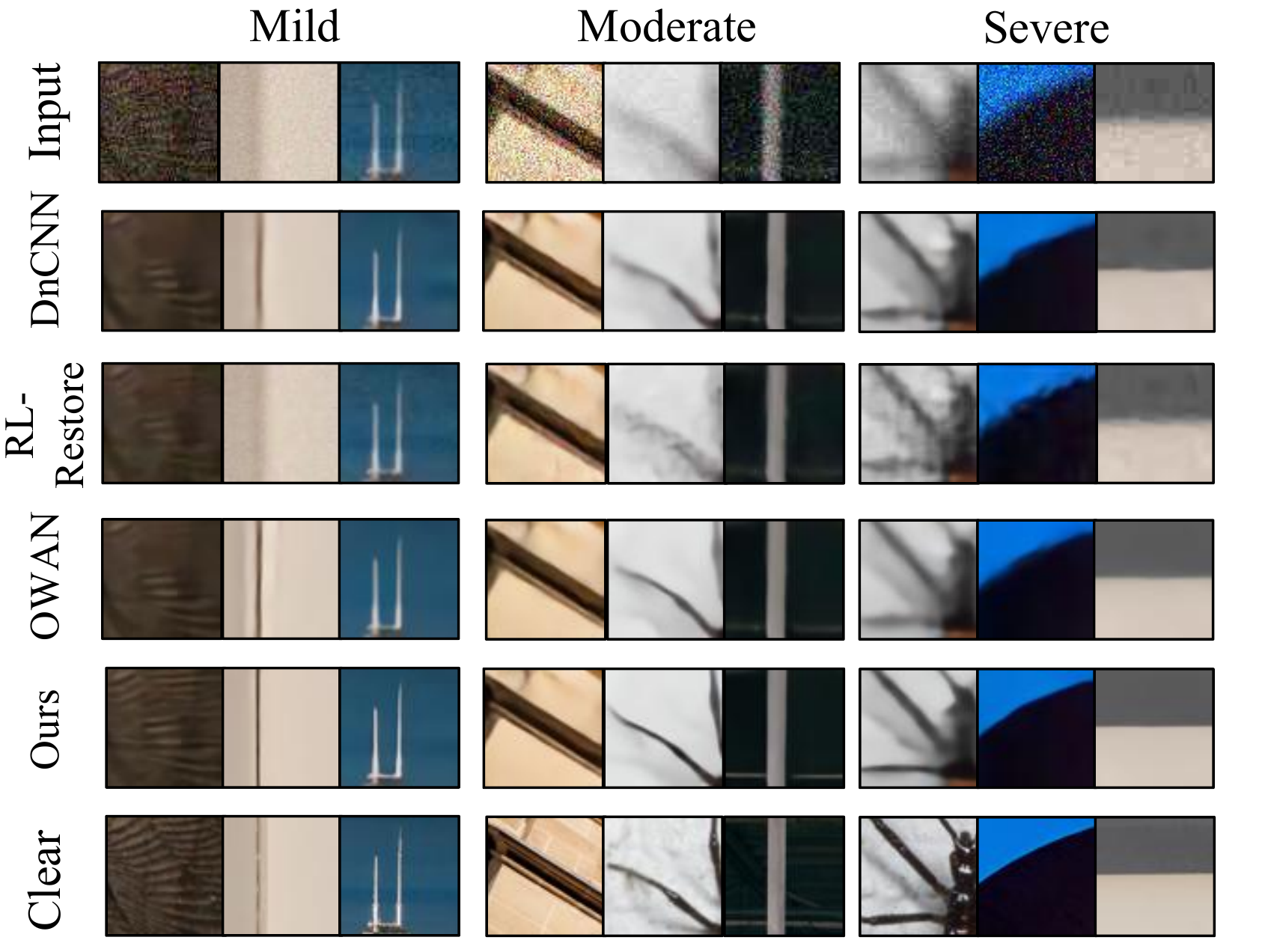} %插入图片，[]中设置图片大小，{}中是图片文件名
\caption{Examples of restored images by our method (H-OWAN-2order), OWAN, RL-Restore, and DnCNN.} %最终文档中希望显示的图片标题
\label{compare} %用于文内引用的标签
\end{figure}

\begin{figure*}[ht] %H为当前位置，!htb为忽略美学标准，htbp为浮动图形
\centering 
\includegraphics[width=0.9\textwidth]{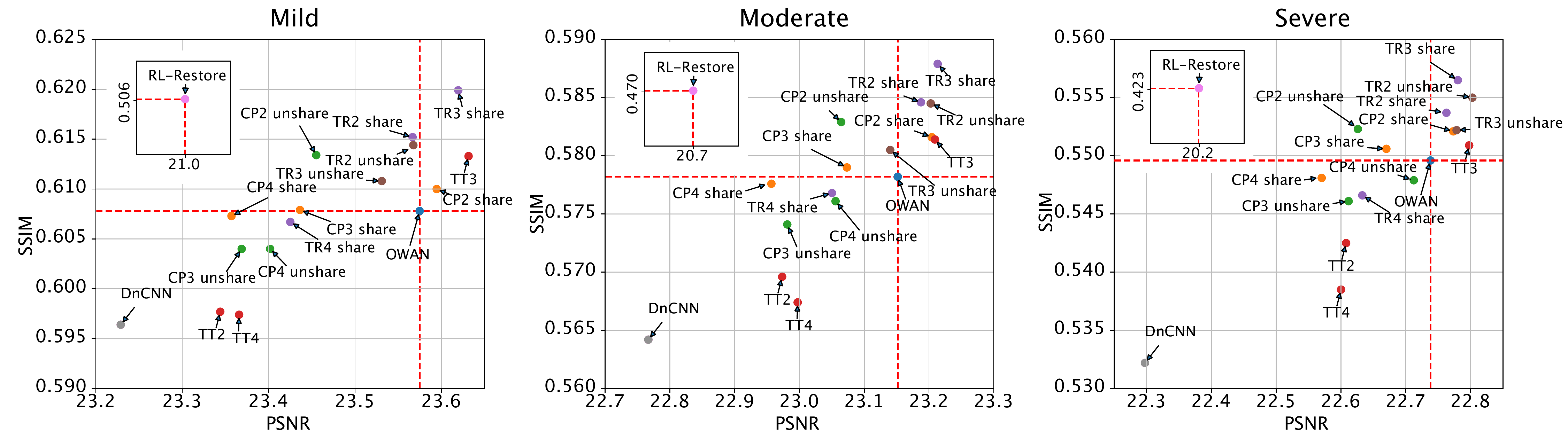} %插入图片，[]中设置图片大小，{}中是图片文件名
 %最终文档中希望显示的图片标题
%  \caption{Experimental results in E3. }
  \caption{PSNR and SSIM results for image restoration with 5 types of distoration with different orders and tensor network decompositions in H-OWAN, where the number of blocks is set as 10.}
 \label{fig:format_compare}
\end{figure*}

% \subsubsection{Comparison with State-of-the-art Methods}

\noindent\textbf{Experimental results}
We compare the performance of H-OWAN with the SOTAs including DnCNN~\cite{zhang2017beyond}, RL-Restore~\cite{yu2018crafting}, Path-Restore~\cite{yu2019path}, OWAN~\cite{suganuma2019attention}.
The experimental results are shown in Table~\ref{Table1:3 noise compare}, where we implement H-OWAN with different orders and also consider the cases that incorporate a bias at the end of the feature map before the tensor product~\cite{kolda2009tensor}.
As shown in Table~\ref{Table1:3 noise compare}, H-OWAN under all conditions outperforms the previous SOTA approaches, and the best results are generally obtained when the order equals $2$. Furthermore, imposing additional bias has no significant performance improvement in this experiment.

\begin{table}[h]\footnotesize

\centering
\begin{tabular}{|c|c|c|c|c|c|}
\hline
VOC                   & Method          & mAP  & VOC                   & Method          & mAP  \\ \hline
\multirow{3}{*}{2007} & w/o Restore & 33.8 & \multirow{3}{*}{2012} & w/o Restore& 32.0 \\
                      & OWAN            & 51.0 &                       & OWAN            & 51.2 \\
                      & Ours            & \textbf{52.3} &              & Ours            & \textbf{52.4} \\ \hline
\end{tabular}

\caption{\textbf{Results on PASCAL VOC.} Comparison of OWAN and our HOWAN-order2. A pretrained SSD300 is applied on distorted images ("w/o" Restore) and their restored versions.}
\label{mAP}
\end{table}

\begin{figure}[h] %H为当前位置，!htb为忽略美学标准，htbp为浮动图形
\centering 
\includegraphics[width=0.45\textwidth]{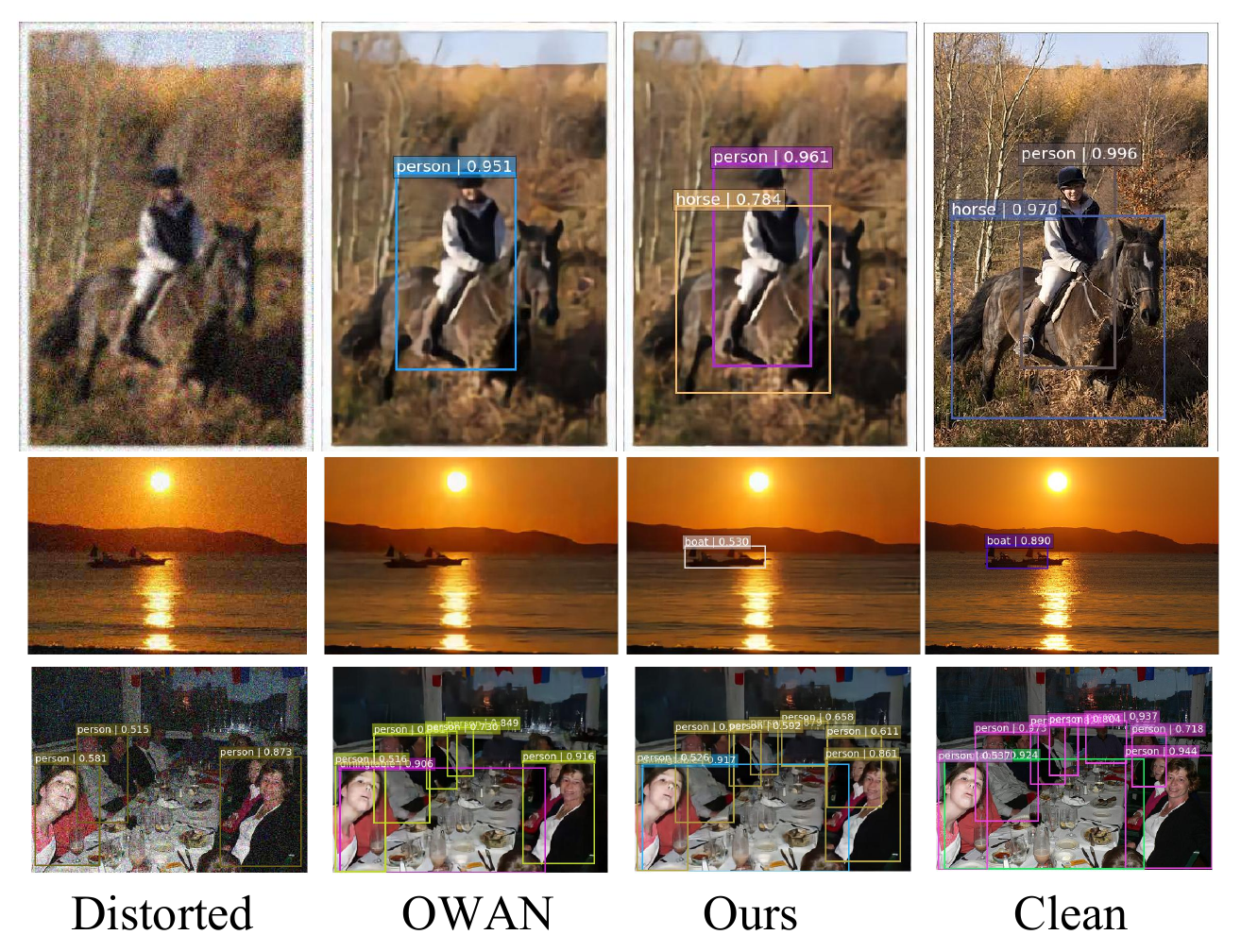} %插入图片，[]中设置图片大小，{}中是图片文件名
\caption{Examples of object detection results on PASCAL VOC, where the color of boxes indicate class categories.} %最终文档中希望显示的图片标题
\label{SSD} %用于文内引用的标签
\end{figure}

% \subsection{Evaluation on Object Detection}
\noindent\textbf{Subsequent objective detection}
The image restoration task is generally employed as a pre-processing module following higher-level computer vision tasks. 
we therefore further evaluate the restoration performance by a subsequent object detection (OD) task, where we use the classic SSD300~\cite{liu2016ssd} and corrupted/restored the PASCAL VOC test set in the experiment.
Table \ref{mAP} shows the mAP results where ``w/o Restore'' denotes ``without restoration'', and Figure~\ref{SSD} gives several illustrative examples of the experimental results. The results can demonstrate the effectiveness of H-OWAN.

% PSNR and SSIM objectively evaluate the image quality, but could not visually reflect the experimental results. Therefore, we refer to the experimental process used in the paper, namely, the denoised images are sent to the object detection network for testing. Here we adopt SSD300 for our experience. The data set we use is PASCAL VOC test set. Three distortions at moderate level (Gaussian noise, Gaussian blur, and JPEG compression) are added to the test set, details see Sec~\ref{subsection421}. 

% Table.\ref{mAP} shows the mAP values, which is obtained by using clear images, mixed noise images, OWAN denoised images and our HOWAN-order2 denoised images as the inputs of SSD300. Fig.~\ref{SSD} shows some detail results on detection. As we can see, the restored images from our method obtains a much higher detection accuracy compared to the distorted images. And our method is more effective than OWAN on this task. 

% \begin{figure}[h] %H为当前位置，!htb为忽略美学标准，htbp为浮动图形
% \centering 
% \includegraphics[width=0.45\textwidth]{figure/only_moderate_3noise.pdf} %插入图片，[]中设置图片大小，{}中是图片文件名
%  %最终文档中希望显示的图片标题
%  \caption{The results of PSNR and SSIM with different orders, number of blocks and rank. In the first line we fix the rank equaling 16, while in the second line we fix the number of blocks equaling 3 for illustration. The red dashed line shows the result by OWAN.}
%  \label{num_compare}
% \vspace{-0.3cm}
% \end{figure}

\begin{figure*}[ht] %H为当前位置，!htb为忽略美学标准，htbp为浮动图形
\centering 
\includegraphics[width=0.87\textwidth]{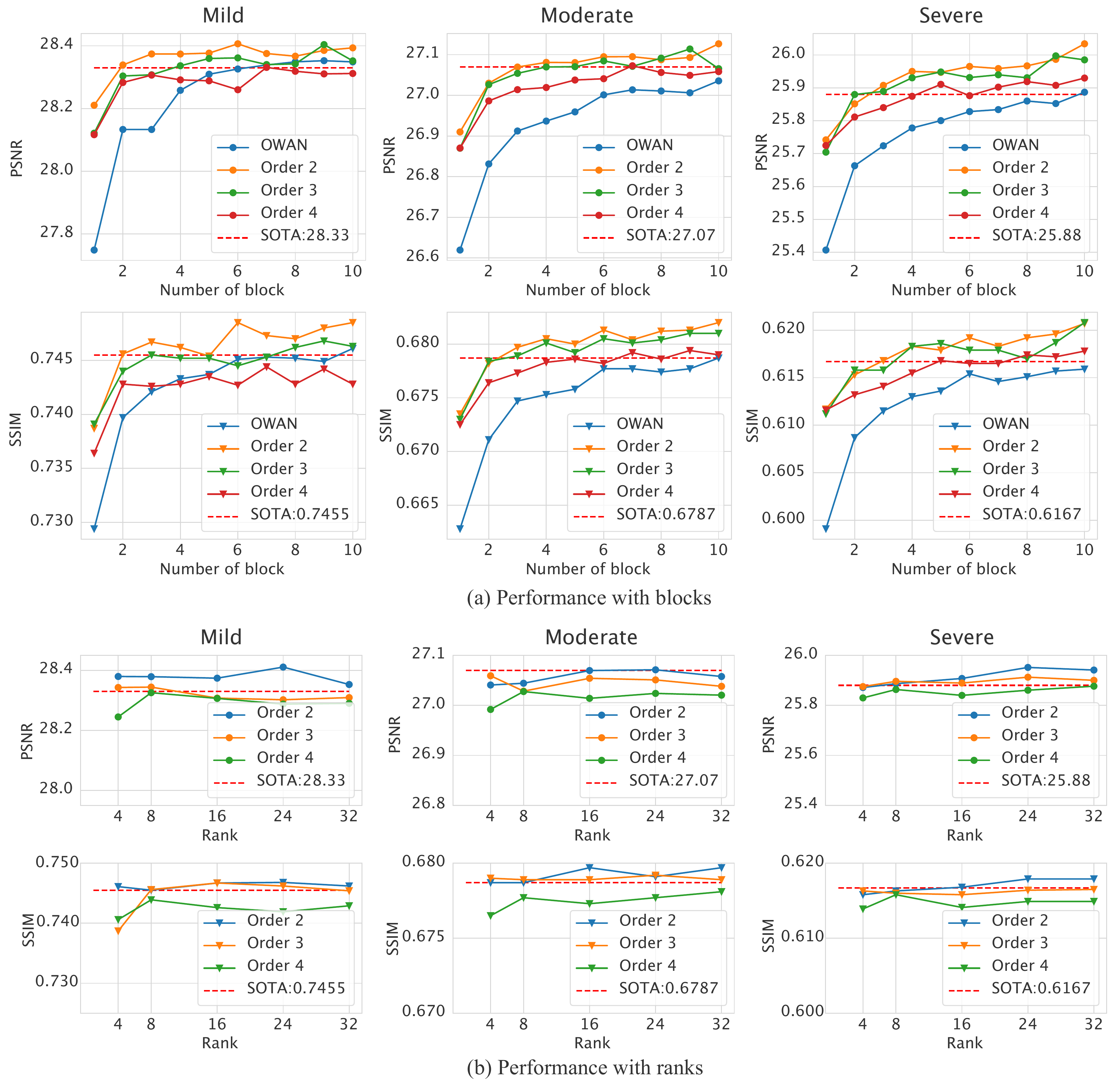} %插入图片，[]中设置图片大小，{}中是图片文件名
 %最终文档中希望显示的图片标题
 \caption{The results of PSNR and SSIM with different orders, number of blocks and rank. In (a) we fix the rank equaling 16, while in (b) we fix the number of blocks equaling 3 for illustration. The red dashed line shows the result by OWAN.}
 \label{num_compare}
\vspace{-0.3cm}
\end{figure*}

\subsection{E2: Ablation Study on Hyperparameters}
% \subsubsection{Tensor $1\times1$ convolution versus $1\times1$ convolution}
% \input{IJCAISections/441ConvolutionCompare}
% \subsubsection{Different number of block, order, and rank}

In this experiment, we evaluate the impact of the additional hyperparameters by T1CLs.
In addition, we also concern whether the performance of the network equipped with T1CLs can be improved with increasing the depth of the network.
Figure~\ref{num_compare} shows the experimental results with all distortion level by (H-)OWAN under various orders, ranks and the number of OWAN blocks.
As shown in Figure~\ref{num_compare} (a), H-OWAN outperforms OWAN under all possible number of blocks and orders.
With increasing the number of blocks, the restoration performance also gradually improves.
However, the performance unexpectedly degenerates with increasing the order.
We infer the reason for such results is because the representation power of order equaling 2 is sufficient for the current task, and higher order would lead to the training difficulty.
The results in the next experiment will show that H-OWAN with higher orders has more promising performance on a more difficult task.
On the other side, the results in Figure~\ref{num_compare} (b) show the performance of H-OWAN is not sensitive with the change of rank of T1CL.

\subsection{E3: IR under Five Types of Distortion}

Below, we use numerical experiments to demonstrate that H-OWAN with higher orders has promising performance on more challenging tasks.

\noindent\textbf{Experiment setting}
Compared to E1, we consider additional two types of distortion, \textit{i.e.} raindrop and salt-and-pepper noise.
The dataset adopted in this experiment is from Raindrop~\cite{qian2018attentive}, where we cropped the original images into 123,968, 5,568 and 25,280 patches for training, validation and test, respectively. 
Apart from CP decomposition, we also represent the tensor kernel in T1CL by TT and TR given in Sec. 3.
Other setting for both the data and networks are same to ones in E1.

\noindent\textbf{Experimental results}
Figure \ref{fig:format_compare} shows the performance constellation including many variants of H-OWAN.
In the figure, the number following specific TN decomposition like ``CP3'' and ``TR4'' denotes the order used in the network, and the keywords ``(un)share'' represents whether assuming the symmetric structure of the kernels in T1CLs.
As shown in Figure \ref{fig:format_compare}, H-OWAN with $3$-order tensor ring format obtains the SOTA performance.
More interestingly, with increasing the strength of the distortion, \textit{i.e.} from mild to severe level, more points appear on the right-top counter of this figure.
It can be inferred that the H-OWANs with higher orders and sophisticated TN decomposition would have more promising performance to handle more challenging restoration tasks.

\section{Conclusion}

Compared to the original OWAN, its high-order extension, \textit{a.k.a.} H-OWAN, achieves the state-of-the-art performance on the multi-distorted image restoration task (see Table~\ref{compare}).
Furthermore, the performance improvement is always kept under various hyper-parameters and configurations (see Figure \ref{num_compare}).
We therefore argue that the proposed tensor $1\times{}1$ convolutional layer (T1CL) not only can effectively alleviate the heterogeneity of features by multiple operations (see Figure \ref{fig:HeteroExpResults}), but also provides powerful representation ability due to the additional non-linearity by tensor product (see Figure \ref{fig:format_compare}).

\clearpage

\small
\bibliographystyle{plain}
\bibliography{ijcai20}

\end{document}